\documentclass[letterpaper]{article} 
\usepackage[preprint]{aaai2027}  
\usepackage[hyphens]{url}  
\usepackage{graphicx} 
\usepackage{natbib}  
\usepackage{caption} 
\usepackage{booktabs}
\usepackage{amsmath}
\usepackage{amssymb}

\title{When Do Internal Probes Beat Reading the Answer?\\Miscalibrated Readouts and Behavior-Concealed Knowledge in Language Models}
\author{
    Gnaneswar Villuri,
    Hashmath Shaik,
    Alex Doboli
}
\affiliations{
    Department of Electrical and Computer Engineering\\
    Stony Brook University\\
    Stony Brook, NY 11794, USA
}

\newcommand{\auc}[1]{#1}

\begin{document}

\maketitle

\begin{abstract}
A 0.6B language model, asked to verify 1{,}200 logical conclusions (half valid, half corrupted by a single semantic edit), answers YES every single time. Judged by behavior it discriminates nothing; linear probes on its hidden states read the correct verdict at \auc{0.96}, transferring to logical structures the probe has never seen and separating foils built from exactly the words of the true conclusion (\auc{0.90}). We ask where the verdict is lost, and find the dominant failure in this controlled setting is a single scalar. The verdict survives to the model's own output logits (margin AUC \auc{0.89}) along a well-aligned readout direction; a saturated decision threshold, offset by $+4.6\sigma$, erases it. This diagnosis is quantitative and it generalizes. Across the 90 semantic-label configurations of a five-model factorial (three families), behavioral accuracy collapses onto a single function of threshold offset (Spearman $-0.93$) while the margin's ranking moves far less. Across a 13$\times$ scale range, internal knowledge saturates while free-form behavior is \emph{non-monotone}: an 8B model underperforms its 4B sibling through an answer-channel failure rather than the threshold; forced-choice accuracy is monotone. The diagnosis is also actionable: a one-parameter correction, never fit on the evaluated structures, repairs behavior from 50\% to 81\% (0.6B), calibrated margin decoding recovers 94\% at 8B by bypassing a free-form channel failure, and few-shot prompting works predominantly the same way, recentering the threshold ($+4.6\sigma \rightarrow 0.0\sigma$) while largely preserving the ranking. Comparing probe to margin further separates three regimes: concealed, miscalibrated, and undetected. On a spatial-reasoning (maze) task built so foils carry no surface cues, the audit correctly reports the third. Finally, in the standard generation setting, answer-surface features and heuristic labels reproduce published probing results without any internal access.
\end{abstract}

\section{Introduction}

\begin{figure*}[t]
\centering
\includegraphics[width=0.98\textwidth]{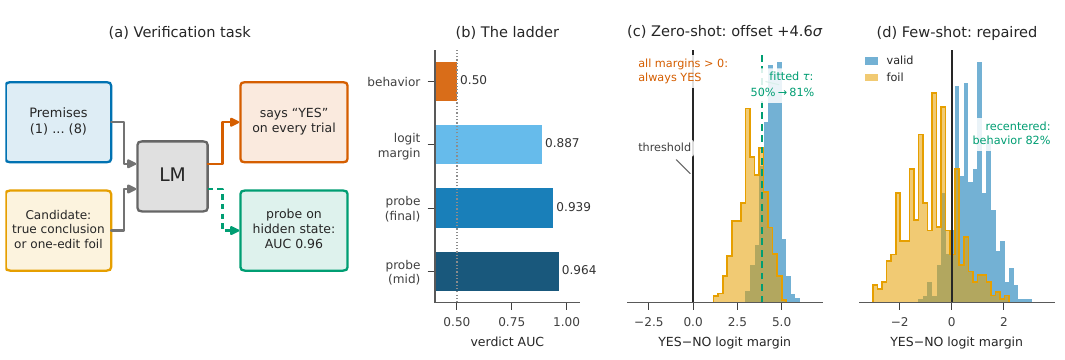}
\caption{The approach and the finding, on real data (Qwen3-0.6B). (a) The model verifies whether a candidate conclusion follows from premises; half the candidates are foils differing by one semantic edit, so labels are exact and the answer is a single token. (b) The verdict is recoverable at every access level, forming a strict ladder; behavior nonetheless sits at chance because the model answers YES on every trial. (c) The mechanism: the YES$-$NO logit margin ranks validity well (valid trials right of foils), but the whole distribution sits $+4.6\sigma$ above the decision threshold, so greedy decoding always says YES; a single fitted threshold $\tau$ repairs behavior to 81\%. (d) Four worked examples perform the same repair: the margin distribution recenters on the threshold while its ranking is unchanged (paired rank correlation 0.79), and behavior recovers to 82\%.}
\label{fig:overview}
\end{figure*}

Language models are trained, in their final and most consequential stage, to produce answers people accept. Human preference signals reward responses that are fluent, confident, and agreeable; nothing in this process directly supervises the model's internal states, which no user ever sees. This asymmetry suggests a specific failure mode: optimization pressure on the \emph{answer channel} could decouple what a model says from what it computes, so that a model sounds sure while its own internal evidence points the other way \citep{sharma2023towards,kadavath2022language}. If so, the internals are the natural place to look: a ``lie detector'' that reads the computation rather than the confession. This paper asks when that picture is measurably true, what breaks between computation and answer, and whether it can be repaired.

A concrete case sets the stakes (Figure~\ref{fig:overview}): a small model judges 1{,}200 times whether a candidate conclusion follows from logical premises; half the candidates are valid, half sabotaged by a single semantic edit. The model answers YES every time; its accuracy is pinned at exactly 50\%. Judged by what it says, it understands nothing. Yet a linear classifier reading its layer-18 activations, trained only on \emph{other} logical structures, separates valid from sabotaged conclusions at \auc{0.96} AUC. The verdict is there to be read. The model just never tells us.

Claims of this general shape (``models know more than they show'') are now common \citep{orgad2025llms,azaria2023internal,kadavath2022language,burns2023discovering}. Our contribution is not the claim; it is locating, quantifying, and repairing the mechanism, under controls strong enough to trust. Doing so connects two literatures that rarely cite each other: the hidden-knowledge claims of interpretability, and the label-prior pathologies documented for prompted classification \citep{zhao2021calibrate,holtzman2021surface,zheng2024large}. Our measurements quantify how much of the former is the latter, and how much is not. Prior work established the dissociation and has begun to intervene on it at the representation level, realigning knowledge and prediction subspaces at inference time \citep{park2025bridging}. We ask a prior question: how much of the gap requires internal access at all? Tracing the verdict through every access level between hidden state and answer, we find it is lost at the last step. The verdict reaches the model's own output logits: the YES$-$NO logit margin, with no probe and no training, ranks validity at \auc{0.887}. The direction the model reads out is well-aligned: at 0.6B it achieves 75\% of the class separation of the best held-out probe direction, and by 4B the two are indistinguishable. What fails at 0.6B is a single scalar: the margin distribution sits entirely on the YES side of the threshold ($+4.6\sigma$), so greedy decoding answers YES regardless of the evidence. The knowledge is not hidden. It is mis-thresholded.

That diagnosis predicts a great deal, and we confirm each prediction. If behavior is ranking passed through an arbitrary threshold, behavior should track the offset wherever it moves. Across the 90 semantic-label configurations of a 120-configuration factorial (five models from three families $\times$ phrasings $\times$ label vocabularies $\times$ option orders), behavioral accuracy collapses onto a single function of measured threshold offset (Spearman $-0.93$). Scale does not make the answer channel reliable either. From 0.6B to 8B, probe-measured knowledge saturates (\auc{0.964} $\rightarrow$ \auc{0.989}) and margin ranking with it (\auc{0.887} $\rightarrow$ \auc{0.980}), and forced-choice accuracy rises monotonically (0.500 $\rightarrow$ 0.923 $\rightarrow$ 0.943); yet free-form accuracy \emph{drops} to 84.8\% at 8B (4B: 93.2\%) through a second channel failure: 113 of 1{,}200 free-form responses contain neither answer token. And if the dominant failure is one number, one number should fix it: a single bias correction, fit by leave-one-structure-out cross-validation and never on the evaluated structure, lifts the 0.6B model from 50.0\% to 81.0\%; at 8B, decoding the margin with one fitted threshold yields 94.2\%, bypassing the free-form failure. Few-shot prompting, we show by paired measurement, performs predominantly this repair: four worked examples move the threshold from $+4.64\sigma$ to $+0.01\sigma$ and behavior to 81.9\%, while largely preserving the margin's ranking of the same trials.

Two further results discipline the claim. First, the probe-margin comparison is itself a diagnostic, and it discriminates. We built a spatial-reasoning task (an ant follows cardinal moves on a grid; the candidate states its final position) in which no foil can be detected from the candidate alone: foil positions are a derangement of the true endpoints, and a candidate-only control measures \auc{0.500}. On paths too long for the model to track silently, the audit finds essentially nothing (margin \auc{0.54}, probe \auc{0.58}): no concealed verdict, correctly reported. On short paths the model partially computes, and the ladder reappears (margin \auc{0.59} $<$ probe \auc{0.70}, gap $+0.118$). Second, we show why this measurement discipline matters: in the standard \emph{generation} setting, where prior probing claims live, a 10-feature baseline over the visible answer text matches or beats hidden-state probes, and heuristic correctness labels manufacture probe signal that survives transfer controls. Without closing the output channel, ``the model knows more than it shows'' cannot be established at all.

\noindent Our contributions:
\begin{itemize}
\item \textbf{The finding: most of what prior work calls ``hidden knowledge'' is already in the output logits.} In a controlled verification testbed with exact labels and structure-held-out evaluation, the model's own YES$-$NO logit margin ranks validity at \auc{0.887}--\auc{0.980} with no probe or training; the probe adds $+0.076$ at 0.6B, shrinking to $+0.008$ at 8B (Tables~\ref{tab:verification},~\ref{tab:ladder},~\ref{tab:crossmodel}). In the standard generation setting, even this margin baseline is unnecessary: a 10-feature answer-surface baseline matches or beats probes in transfer, and heuristic labels manufacture probe signal that survives transfer controls.
\item \textbf{The mechanism: the failure is a threshold offset, not a direction error.} The readout direction is well-aligned (optimal by 4B); what fails is a single scalar ($+4.6\sigma$). Across 90 prompt configurations of five models (three families), behavior collapses onto one function of this offset (Spearman $-0.93$; Figure~\ref{fig:factorial}) while the ranking barely moves. Scale saturates internals by 4B but does not make the output channel reliable.
\item \textbf{The fix: one parameter repairs the gap, the diagnostic says when.} Bias-only calibration recovers 50\%$\rightarrow$81\% (0.6B) and 85\%$\rightarrow$94\% (8B), matched by median-centering; few-shot prompting performs predominantly this same recentering ($+4.64\sigma$$\rightarrow$$+0.01\sigma$, rank correlation 0.79). A probe-vs.-margin comparison further tells you \emph{which} failure you have: on a spatial-reasoning task immune to surface cues (candidate-only AUC $= 0.500$), the diagnostic correctly reports absence when the model cannot solve the task and a miniature ladder when it partially can.
\end{itemize}

\section{Related Work}

\paragraph{Error detection from internal states.}
\citet{kadavath2022language} showed models can estimate the probability their own answers are correct; \citet{azaria2023internal} trained classifiers on activations to detect false statements; \citet{burns2023discovering} recovered truth directions without supervision; \citet{marks2024geometry} mapped the linear geometry of true/false representations; \citet{li2023inference} located and intervened on truthfulness-relevant heads. Closest to us, \citet{orgad2025llms} probed exact-answer tokens and reported that models sometimes encode correct answers while generating wrong ones; we adopt their token-position and layer methodology. \citet{gekhman2025inside} formalize the gap directly, scoring answer candidates from hidden states versus observable token probabilities and measuring a 40\% relative internal--external gap on factual QA. Closest in aim, \citet{park2025bridging} intervene on the gap at the representation level: locating distinct knowledge and prediction subspaces in the residual stream and realigning them at inference time on multiple-choice benchmarks. Our diagnosis differs: under a closed-channel verification design with exact labels and output-surface controls, most of the gap is already present in the model's own output margin, dominated by a scalar threshold offset, and repairable with no internal access at all. We also mark where that account ends (symbol grounding, computation failure) and why open-channel designs cannot establish it. Localization to the readout, the answer-surface control, and the margin baseline are the steps prior work leaves open.

\paragraph{Probing methodology.}
The probing literature warns that decodability is not use \citep{belinkov2022probing} and that probe accuracy needs controls to separate representation signal from probe capacity \citep{hewitt2019designing}. Our controls extend this tradition in the direction correctness-probing requires: the null is not a shuffled task but \emph{the information available without internal access}: answer text, task difficulty, a frozen encoder \citep{xiao2023cpack}, and the model's own logit margin, read through the unembedding as in the logit-lens family \citep{nostalgebraist2020logit,belrose2023eliciting}. In vision-language models, first-token logit distributions likewise carry decision-relevant signal that later tokens lose \citep{zhao2024first}. Output-distribution methods for hallucination detection \citep{farquhar2024detecting,kossen2024semantic} occupy one rung of the resulting ladder; our measurements bound what such methods can capture in principle, and show the bound is nearly reached.

\paragraph{Label priors and calibration.}
That prompted classification is distorted by label priors is established: few-shot predictions shift with example order and label frequency, repaired by an affine correction fit on content-free inputs \citep{zhao2021calibrate}; surface-form competition depresses valid answers \citep{holtzman2021surface}; multiple-choice selections carry systematic position bias, removable by prior estimation \citep{zheng2024large}; label bias persists across hundreds of tasks even after debiasing \citep{reif2024beyond}; miscalibrated confidence is classical \citep{guo2017calibration}; and preference tuning induces agreement biases \citep{sharma2023towards}. This literature repairs the output distribution of a black box; we open the box and measure what the prior does to information already computed. The ladder shows the verdict is present internally at near-ceiling, aligned with the readout direction, and quantifies how much of the probing literature's ``hidden knowledge'' is exactly this known pathology (most of it) versus concealment beyond any output-distribution repair ($+0.008$--$+0.076$ AUC). The diagnostic also marks where the calibration account \emph{ends}: symbolic-label inversions are sign errors no bias correction can repair, and extended reasoning exceeds the margin's ceiling. Few-shot prompting \citep{brown2020language} and chain-of-thought \citep{wei2022chain} are standard elicitation tools; we give a mechanistic account of what the former does in this setting.

\section{A Controlled Testbed}

\paragraph{Corpus and verification task.}
We use the parallel logic corpus of \citet{zhou2026geometry}, who study representation trajectories of models \emph{reading} these proofs without relating them to correctness; we repurpose it into verification tasks with exact labels. The corpus has 30 natural-deduction structures (8--16 steps), each instantiated in 20 topical domains with aligned steps ($30\times20=600$ English instances). The model receives an instance's premises and a candidate conclusion and must answer in one word whether the candidate follows. The core prompt is fixed verbatim: premises as a numbered list, then ``Candidate conclusion: [candidate]'', then ``Does this conclusion follow from the premises by logical deduction? Answer with exactly one word: YES or NO.'' (chat template applied; two variants in the factorial). Each instance yields a gold-YES trial (the true conclusion) and gold-NO foils. Labels are exact by construction, the answer surface is a single constant-format token, and because every structure appears in every domain, structure- and domain-identity shortcuts are detectable by holding out entire structures.

\paragraph{Foils.}
Two generations of foils guard against generator artifacts. \emph{Minimal foils} ($n{=}600$; one per instance) differ from the true conclusion by exactly one semantic edit: polarity flip ($n{=}527$), entity swap using premise entities ($n{=}46$), or conjunct swap exchanging constituents \emph{across} the two conjuncts, producing a non-entailed proposition from the same words (``the \emph{application} builds successfully and CI marks the \emph{module} as tested'' $\rightarrow$ ``the \emph{module} builds successfully and CI marks the \emph{application} as tested''; $n{=}27$; 24 preserve the word multiset exactly). They were produced by an LLM under a constrained rubric and independently checked (non-entailment, single edit, grammaticality). \emph{Deterministic foils} ($n{=}503$) are generated by four rules with no LLM anywhere: negation of the conclusion's main clause (provably non-entailed given consistent premises; $n{=}329$), unsafe quantifier strengthening $\exists\rightarrow\forall$ ($n{=}106$), role swaps that permute argument structure under verb-agreement constraints, preserving the word multiset ($n{=}46$), and entity swaps ($n{=}22$). The balanced core benchmark is $600+600=1{,}200$ trials; the expanded set adds the 503 deterministic foils.

\paragraph{Spatial verification (maze).}
Because logic foils are sentences, a foil could in principle be recognized by \emph{plausibility} of the candidate alone. Our second domain eliminates this channel by construction: an ant starts at the origin and follows a sequence of cardinal moves (``move 3 units left; move 2 units up; \ldots''); the candidate states its final position. A true candidate (``the ant is at $(-3, 2)$'') and a foil (``the ant is at $(1, -1)$'') are formally indistinguishable coordinate pairs. Foil positions are a \emph{derangement} of the true final positions within each step-count group: every foil is another chain's true endpoint, so the position distributions are identical by construction. A candidate-only probe confirms this (\auc{0.500} across all tiers and models). We use a hard tier (4--6 steps, 1{,}600 trials) and an easy tier (2--3 steps, small moves, 1{,}300 trials), with transfer evaluated across step-count groups.

\paragraph{Probing protocol and statistics.}
Probes are $\ell_2$-regularized logistic regression on standardized single-layer, single-position hidden states (answer-token and last-input-token positions), following \citet{orgad2025llms}. The primary evaluation is leave-one-structure-out (LOSO): train on 29 structures, evaluate on the held-out one, cycle. To eliminate layer-selection optimism, the probe layer is chosen \emph{inside} each fold by inner cross-validation over training structures only (``nested''); the selection is stable (all 30 folds at 0.6B choose layer 18) and nested matches non-nested to $\pm0.001$. All confidence intervals are \emph{cluster} bootstraps resampling structures, not trials (2{,}000 resamples), since trials within a structure are correlated; significance tests are label permutations within structures. Exact prompts, templates, checkpoints, decoding and scoring rules, folds, and foil-generation code are in the supplement. Models: Qwen3-0.6B/1.7B/4B/8B \citep{yang2025qwen3}, Phi-3.5-mini (3.8B; \citealt{abdin2024phi}), and SmolLM2-1.7B \citep{allal2025smollm2}, plus the Qwen3-0.6B/4B base (pre-instruction-tuning) checkpoints for the training analysis in the Discussion.

\section{Knowledge, Expression, and a Diagnostic}

We phrase our terms with usable information under computational constraints ($\mathcal{V}$-information; \citealt{xu2020theory}), which is not vacuous for deterministic networks. Let $h_\ell(x)$ be the layer-$\ell$ state on trial $x$, $V(x)\in\{0,1\}$ the validity label, $A(x)$ the emitted answer. The model \emph{knows} $V$ at layer $\ell$ to the degree a linear observer predicts $V$ from $h_\ell$; it \emph{expresses} $V$ to the degree $V$ is predictable from $A$. We estimate both with AUC.

\paragraph{Proposition 1 (readout bottleneck).}
For single-token verification under \emph{forced-choice} decoding (the decoder restricted to the two answer tokens), the answer is a deterministic function of the scalar margin $m(x)=u^\top h_L(x)$, where $u$ is the difference of unembedding rows for the two answer tokens and $h_L$ the final post-norm state: $A=\mathrm{sign}(m)$. Free-form generation can deviate from this object, including by emitting neither token; we report the two behaviors separately. The chain $h_L\rightarrow m\rightarrow A$ is a sequence of garblings, so by the data-processing property of statistical experiments \citep{blackwell1953equivalent} the best achievable prediction of $V$ under any proper loss cannot improve from left to right. Expression is the special case: one particular direction must align with the validity feature \emph{and} its threshold must fall inside the data. Concealment is generic.

\paragraph{Proposition 2 (three failures, one diagnostic).}
Suppose behavior is uninformative ($\mathrm{AUC}(A)\approx0.5$). Three regimes are distinguishable from two further measurements. (i) \emph{Miscalibration}: $\mathrm{AUC}(m)$ high; the margin ranks validity but an offset places every margin on one side of the threshold. (ii) \emph{Misalignment (concealment)}: probe AUC high but $\mathrm{AUC}(m)\approx0.5$; the verdict is computed but the readout direction misses it. (iii) \emph{Absence}: probe AUC $\approx \mathrm{AUC}(m) \approx$ low; no additional verdict is linearly decodable at any layer, so there is nothing for internal access to recover. The regimes have different consequences: (i) predicts one-parameter repair and repair-by-recalibrating-elicitation; (ii) predicts neither works without internal access; (iii) predicts nothing works.

\section{Verification: Knowledge Without Behavior}

\begin{table}[t]
\centering
\small
\begin{tabular}{@{}lc@{}}
\toprule
Measurement (Qwen3-0.6B) & Value \\
\midrule
Behavior, balanced 1{,}200 trials (all YES) & 0.500 (exact) \\
Behavior, 503 deterministic foils (all YES) & 0/503 \\
\quad options reordered / question negated & all YES / 0.38 \\
\midrule
Probe, nested LOSO (layer 18 in all 30 folds) & 0.964 [.950, .976] \\
Probe accuracy at threshold 0.5 (LOSO) & 0.898 \\
Permutation test (within-structure) & $p \le 10^{-3}$ \\
Per-structure AUC & all $\ge 0.90$ \\
\midrule
Expanded foil set (1{,}703 trials), LOSO AUC & 0.968 \\
\quad negation / polarity / quantifier & 0.994 / 0.985 / 0.975 \\
\quad role swap / conjunct swap / entity swap & 0.917 / 0.864 / 0.76--0.81 \\
\quad macro-average over families & 0.900 \\
\quad identical-word foils combined ($n{=}73$) & 0.897 \\
\midrule
Lexical overlap baseline & 0.759 [.725, .797] \\
Frozen text encoder (768-d) & 0.642 \\
Cross-family probe, same trials (Phi-3.5) & 0.990 \\
\bottomrule
\end{tabular}
\caption{Verification, Qwen3-0.6B. Behavior carries zero information (YES to all 1{,}703 trials, including the 503 deterministic foils), while structure-held-out probes on its states classify validity across every foil family, far above text-only bounds. Layer selection is nested inside folds; CIs are cluster bootstraps over structures.}
\label{tab:verification}
\end{table}

\begin{figure}[t]
\centering
\includegraphics[width=0.98\columnwidth]{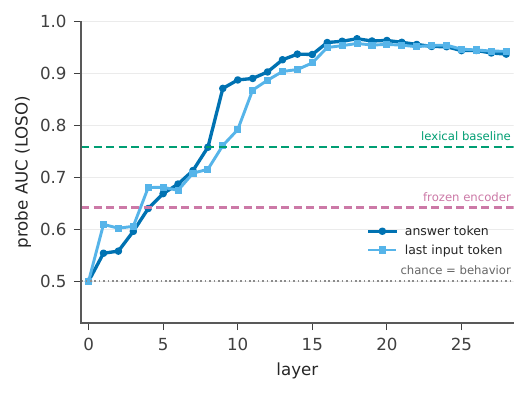}
\caption{The validity signal is computed, not read off the surface: structure-held-out probe AUC by layer (Qwen3-0.6B) at two token positions. Early layers carry little signal (0.50--0.60 through layer 4); it assembles through the middle of the network and plateaus at 0.95--0.96 by layer 16, far above the lexical baseline (0.76) and frozen-encoder (0.64) references. Behavior sits at chance throughout.}
\label{fig:layers}
\end{figure}

\paragraph{Behavior is pinned at chance.}
The 0.6B model answers YES on all 1{,}200 balanced trials and all 503 deterministic foils; balanced accuracy is exactly 50\%. This is not a parsing or position artifact: reordering the options leaves every answer YES, and negating the question (which inverts every gold label) drops accuracy \emph{below} chance to 38\%, the few NO answers concentrating on foils (32/40), a sign-inverted leak of content through a content-blind affirmation.

\paragraph{The internals classify what the behavior cannot.}
A nested LOSO probe on answer-token states reaches \auc{0.964} [0.950, 0.976] and 89.8\% accuracy on the same trials the model scores 50.0\% (Table~\ref{tab:verification}). Every one of the 30 held-out structures probes at $\ge$\auc{0.90}. On the expanded foil set the probe separates all seven foil families, including rule-generated foils no LLM produced: negations at \auc{0.994}, quantifier strengthenings at \auc{0.975}, and, critically, foils sharing their entire word multiset with the truth at \auc{0.897} ($n{=}73$; 70 word-for-word anagrams, the rest differing only in agreement forms). The macro-average over families is \auc{0.900}; the weakest family (entity swap, \auc{0.76}--\auc{0.81}) marks the floor. Text-only bounds fall far short: candidate--premise overlap reaches \auc{0.759}, a frozen general-purpose encoder \auc{0.642}. The layer profile (Figure~\ref{fig:layers}) shows the signal is weak in early layers (at most \auc{0.60} through layer 4) and assembles through the middle of the network: the signature of a computed judgment, not a surface read. Nor is the judgment idiosyncratic: a probe trained on Qwen3-0.6B structures and applied to \emph{Phi-3.5's} states reads validity at \auc{0.990}, so the computation is model-general.

\paragraph{How much is verification, how much plausibility?}
A candidate might still be separable by semantic plausibility alone, without the premises. We measure this directly with two ablated probes on the same states: \emph{candidate-only} (premises removed from the input) and \emph{premise-shuffled} (premises replaced by another instance's). Overall, the matched probe reaches \auc{0.962} against \auc{0.817} candidate-only and \auc{0.828} shuffled. A genuine premise-dependent component exists, but much of the aggregate signal is candidate-intrinsic: negating a sentence leaves plausibility fingerprints. On role swaps, where the foil is a grammatical sentence using identical words, candidate-only collapses to \auc{0.592} while the matched probe holds \auc{0.915}: a $+0.32$ premise gap. On negations, candidate-only alone reaches \auc{0.933}. The probe reads \emph{both} channels; the premise-dependent one is largest exactly where plausibility is silenced by construction, and the maze domain closes that channel entirely.

\section{Where the Verdict Is Lost}

\begin{table}[t]
\centering
\small
\begin{tabular}{@{}l@{\hspace{6pt}}c@{\hspace{6pt}}c@{\hspace{6pt}}c@{}}
\toprule
 & 0.6B & 4B & 8B \\
\midrule
Free-form (acc.) & .500 & .932 & .848 \\
Forced-choice (acc.) & .500 & .923 & .943 \\
Margin (AUC) & .887 & .979 & .980 \\
\quad margins${>}0$ & 100\% & 46.5\% & 53.1\% \\
Probe, LOSO (AUC) & .964 & .989 & .989 \\
\midrule
probe$-$margin & +.076 & +.010 & +.008 \\
\quad 95\% CI & [.057,.096] & [.001,.019] & [.000,.019] \\
\midrule
+ calibration (acc.) & .810 & .927 & .942 \\
\bottomrule
\end{tabular}
\caption{The information ladder across scale (verification, balanced set; cluster-bootstrap CIs over structures). Forced-choice accuracy is monotone; free-form is not (113 of 8B's responses contain neither token). Calibration repairs 0.6B to 81.0\% and 8B to 94.2\% (margin-decoded); 4B gains nothing (92.7\%). 0.6B probes the answer token (layer 18); 4B/8B the last input token (layer 0 = 0.500, confirming no leak).}
\label{tab:ladder}
\end{table}

\paragraph{The ladder is strict, and the failure is late.}
Table~\ref{tab:ladder} traces the verdict through every access level, with paired cluster-bootstrap CIs. At 0.6B: text \auc{0.500} $<$ margin \auc{0.887} [0.864, 0.912] $<$ probe \auc{0.964} [0.950, 0.976]; the probe$-$margin gap is $+0.076$ [$+0.057$, $+0.096$] and no resample of 2{,}000 reverses any rung. The verdict thus survives to the output distribution; regime (ii) concealment is real but small. The dominant failure is regime (i): the margin is positive on 100\% of trials, its mean offset $+4.6\sigma$ above the decision threshold, so the argmax answers YES regardless of the evidence.

\paragraph{The readout direction is well-aligned, and optimal by 4B.}
A direct measurement shows how much the \emph{direction} contributes. Against the strongest comparator, the structure-held-out probe direction itself, the class separation ($d'$) along the model's readout axis is 1.72 vs.\ 2.28 at 0.6B (75\%, matching the $+0.076$ AUC gap) and 3.74 vs.\ 3.74 at 4B. The alignment $\cos(\Delta\mu, u)$ grows from 0.19 to 0.26 with scale. Direction error is real but secondary at 0.6B and gone by 4B; the threshold, by contrast, is set by the prompt rather than by competence.

\begin{figure}[t]
\centering
\includegraphics[width=0.98\columnwidth]{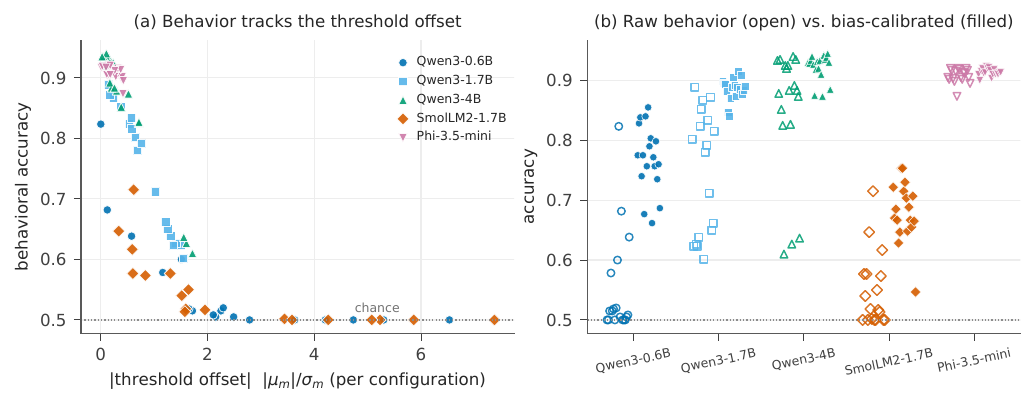}
\caption{Behavior is ranking plus an uncontrolled offset. (a) Across 90 semantic-label prompt configurations of five models (three families), behavioral accuracy is a single declining function of the measured threshold offset $|\mu_m|/\sigma_m$ (Spearman $-0.93$); beyond ${\sim}2\sigma$, behavior is chance regardless of competence. (b) The same configurations after a one-parameter bias correction (filled): per-model spread collapses (e.g.\ 4B: SD 0.107 $\rightarrow$ 0.022), leaving capability differences, not prompt sensitivity.}
\label{fig:factorial}
\end{figure}

\paragraph{Behavior tracks the threshold wherever it moves.}
Proposition 1 makes the threshold account \emph{available}, not true: behavior could vary across prompts because the underlying ranking varies. The factorial's empirical content is that it does not: the offset term, not the ranking, carries the sensitivity. We ran a factorial over five models $\times$ three phrasings $\times$ four answer-label vocabularies $\times$ two option orders (120 configurations). Figure~\ref{fig:factorial} shows the result for the 90 semantic-label configurations: behavioral accuracy collapses onto one declining function of measured offset (Spearman $-0.93$). The relation is not circular: a ranking-driven account would move margin AUC with behavior, and it does not---per-model margin-AUC spread is $2$--$8\times$ smaller than behavioral spread (Phi-3.5: behavior 0.87--0.92, margin \auc{0.954}--\auc{0.968} across its 18 configurations). After the one-parameter correction, per-model spread collapses (4B: SD 0.107 $\rightarrow$ 0.022) and what remains is capability. The remaining 30 configurations mark a boundary. With arbitrary symbol labels (answer ``B'' for supported, ``A'' otherwise), larger models' margins \emph{invert}: 4B reaches margin AUC \auc{0.03} with one option order, near-perfectly wrong. The verdict is intact but its binding to arbitrary symbols sign-flips: a grounding failure, not calibration, and bias correction correctly cannot repair it (calibrated accuracy $\approx$ 0.50).

\paragraph{Scale saturates knowledge; the answer channel keeps failing, differently.}
Across 0.6B $\rightarrow$ 4B $\rightarrow$ 8B, probes move \auc{0.964} $\rightarrow$ \auc{0.989} $\rightarrow$ \auc{0.989}, margins \auc{0.887} $\rightarrow$ \auc{0.979} $\rightarrow$ \auc{0.980}, and forced-choice accuracy 0.500 $\rightarrow$ 0.923 $\rightarrow$ 0.943: by 4B the internal verdict is at ceiling, present in the output distribution, and correctly thresholded. Free-form behavior does not follow: 0.500 $\rightarrow$ 0.932 $\rightarrow$ \textbf{0.848}. The 8B drop is not the threshold returning: 113 of 1{,}200 free-form responses contain neither answer token (accuracy on the parseable remainder is 93.7\%), a second, distinct failure of the emitted channel, instruction drift rather than saturation. Reading the margin with one fitted threshold yields 94.2\%, bypassing both failure modes: scale fixed the verdict and the threshold, not the reliability of free-form text.

\section{Repair, Elicitation, and the Diagnostic}

\begin{table}[t]
\centering
\small
\begin{tabular}{@{}lcc@{}}
\toprule
Condition (Qwen3-0.6B) & Behavior & YES/NO \\
\midrule
Zero-shot & 0.500 & 1200 / 0 \\
+ bias-only calibration (1 param., LOSO) & 0.810 & (margin) \\
+ 4 worked examples & 0.814 & 564 / 556 \\
+ extended reasoning (2{,}048 tokens) & 0.902 & 701 / 495 \\
\midrule
Qwen3-4B, zero-shot & 0.932 & 618 / 582 \\
Qwen3-8B, zero-shot / + calibration & 0.848 / 0.942 & 614 / 473 \\
Phi-3.5-mini, zero-shot & 0.921 & 603 / 597 \\
\midrule
Probe accuracy, zero-shot (0.6B / 4B) & \multicolumn{2}{c}{0.898 / 0.950} \\
\bottomrule
\end{tabular}
\caption{Expression is the moving part. A single bias parameter matches four worked examples (81.0\% vs.\ 81.4\%), and a paired measurement shows they are the same intervention. Extended reasoning and scale approach, and at 0.6B statistically match (0.902 vs.\ 0.898), what zero-shot probes read from the states. Few-shot evaluates 1{,}120 trials on 28 structures (examples from the other two); calibration is evaluated strictly on held-out structures. The 8B calibrated figure (0.942) is margin-decoded, bypassing the 113 invalid free-form responses entirely; the 0.848 zero-shot figure scores them as incorrect.}
\label{tab:repair}
\end{table}

\paragraph{One parameter, cross-validated.}
We fit a single scalar $\tau$ to the margin, and nothing else, on 29 structures and evaluate on the 30th, cycling (LOSO). At 0.6B this lifts behavior from 50.0\% to 81.0\% ($\tau = 2.93 \pm 0.02$ across folds; Platt scaling does no better); at 8B, calibrated margin decoding yields 94.2\%. Median-centering, using only the known base rate, matches the supervised scalar exactly (81.0\%; 94.7\% at 8B), so the repair requires no learning. Content-free calibration \citep{zhao2021calibrate} \emph{fails} (55.9\%; estimates swing $2\sigma$ across placeholders): the offset is anchored to real content, unlike multiple-choice label priors. What is new is not the fix but what its success measures (the distance between behavior and the margin's own ceiling) and its scope: $\tau$ transfers across held-out structures because the offset is global within a prompt, but only partially across phrasings (0.77--0.83 where offsets are similar; 0.63--0.75 where they differ by $2\sigma$), exactly as the factorial's offset variation predicts. The ranking transfers; the offset is re-estimated per prompt configuration.

\paragraph{Few-shot prompting predominantly recenters the threshold.}
The mechanism predicted in advance that worked examples would repair behavior by moving the threshold, not the ranking. We measured margins per trial, zero-shot and with four examples, paired. Zero-shot: AUC \auc{0.880}, offset $+4.64\sigma$, 100\% of margins positive, behavior 50.0\%. Few-shot: AUC \auc{0.893}, offset $+0.01\sigma$ (dead center), 52.2\% positive, behavior 81.9\% under first-token scoring (81.4\% under Table~\ref{tab:repair}'s free-generation scoring; the rules differ on 6 of 1{,}120 trials). The ranking is largely preserved (rank correlation 0.79); the threshold moved $4.6\sigma$. Three independent routes land on the same number because they predominantly perform the same recentering: statistical calibration (81.0\%), worked examples (81.4\%), and the margin's own ceiling at the optimal threshold. Extended reasoning goes further (90.2\%), consistent with reasoning tokens adding computation rather than merely recalibrating; we leave its mechanism open.

\paragraph{The residual gap: internal access still buys errors back.}
On the 82 trials the 4B model answers incorrectly, a probe trained only on other structures recovers the correct verdict at 63.4\% (AUC \auc{0.771}); overall it scores 95.0\% against the model's 93.2\%. The probe$-$margin gap at 4B/8B ($+0.010$, 95\% CI $[+0.0012, +0.0191]$ / $+0.008$, $[+0.0001, +0.0189]$; $P(\Delta \le 0) = 0.012$ and $0.025$) is small but excludes zero. Concealment shrinks with scale; it has not vanished.

\paragraph{Discriminant validity: the audit tracks what is computed.}
On the maze task the 0.6B model again answers YES to every trial at both tiers, replicating the behavioral collapse in a second domain; the audit's other rungs discriminate. On the hard tier (4--6 steps) every rung is near-empty: margin \auc{0.538}, probe \auc{0.581}, candidate-only \auc{0.500}. That is regime (iii): no verdict is linearly decodable beyond what the logits show, and the probe, given 1{,}600 trials, correctly reports nothing to find. On the easy tier (2--3 steps) the model partially computes, and the ladder reappears: margin \auc{0.587} $<$ probe \auc{0.704} (gap $+0.118$), candidate-only \auc{0.500}, and the verdict peaks at layer 18, the same mid-network position as the logic verdict. At 1.7B the calibration story replicates (behavior 52.5\%, margin \auc{0.799}, repaired to 70.9\%; candidate-only \auc{0.500}). The audit therefore does not manufacture gaps: it reports absence where nothing is detectable, a miniature ladder where computation is partial, and the full dissociation where a strong verdict signal forms early.

\section{Generation: Why the Controls Matter}

\begin{table}[t]
\centering
\footnotesize
\begin{tabular}{@{}l@{\hspace{4pt}}c@{\hspace{4pt}}c@{\hspace{4pt}}c@{\hspace{4pt}}c@{\hspace{4pt}}c@{}}
\toprule
 & \multicolumn{3}{c}{Qwen3} & Phi & SmolLM2 \\
\cmidrule(lr){2-4} \cmidrule(lr){5-5} \cmidrule(lr){6-6}
 & 0.6B & 4B & 8B & 3.8B & 1.7B \\
\midrule
Behavior & .500 & .932 & .848 & .921 & .50--.71$^*$ \\
Margin & .887 & .979 & .980 & .963 & .60--.82$^*$ \\
Probe & .964 & .989 & .989 & .990 & --- \\
\midrule
Calibrated & .810 & .927 & .942 & .921$^\dagger$ & --- \\
\bottomrule
\end{tabular}
\caption{The ladder across models and families. The behavior $<$ margin $<$ probe ordering holds in every model where probes were fit; behavior and margin rankings are available for all five. SmolLM2 ($^*$) ranges are across prompt configurations from the factorial (no hidden-state extraction). Phi-3.5 probe is cross-family: trained on Qwen3-0.6B structures, evaluated on Phi states. $^\dagger$Phi margin is already centered (offset $-0.5\sigma$); calibration matches behavior.}
\label{tab:crossmodel}
\end{table}

The verification design came from an audit of the standard generation setting, itself a result. Asked to \emph{write} conclusions (labels by two independent frontier-LLM judges, three-way rubric, 99--100\% agreement), probes on the writing model's states appear to predict correctness, until the controls arrive. A 10-feature baseline over the visible answer (length, format, completeness statistics) matches probes with heuristic keyword labels (transfer AUC 0.870 vs.\ 0.856) and beats them under adjudicated labels at 4B by 12--13 points (0.947/0.970 surface vs.\ 0.818/0.837 hidden state, strict/lenient). Combining features adds nothing: the reason is mundane, since these models fail by writing fragments and half-conclusions, so correctness is worn on the answer's face.

Label quality matters as much as the baseline. Heuristic labels \emph{manufacture} internal signal: keyword-overlap ``correctness'' is partly a function of answer surface, so any surface-correlated representation predicts it, and the artifact survives transfer; adjudication (strict correctness 15.7\% $\rightarrow$ 5.5\%) changes which predictors work. At the pre-answer position, a 4-feature difficulty baseline (premise count, depth, prompt length) beats every hidden-state probe under transfer. None of this shows generation internals are empty; it shows the standard design cannot demonstrate otherwise, because an open output channel explains probe success. Hence: close the channel, then measure.

\section{Discussion and Conclusion}

\paragraph{For interpretability.}
``Models know more than they show'' is, in the cleanest case we can construct, true, and mostly mislabeled. The verdict is not buried in a hidden subspace awaiting a probe; it sits in the output logits behind a broken threshold. The remedies differ (internal access vs.\ one number), and so do the implications: at the scales we test, log-probability access recovers nearly everything, and the concealed remainder ($+0.008$--$+0.076$ AUC) is real but thin. Probing claims should ship with margin and prior-corrected baselines \citep{zhao2021calibrate}; where probes beat both, \emph{that} is the discovery.

\paragraph{For evaluation.}
Behavioral accuracy on balanced verification tasks can sit arbitrarily far below competence, and the direction of error is not fixed: the same model saturates to all-YES or all-NO by prompt wording. Text-only evaluation measures ranking $\times$ threshold, and the threshold term dominates prompt sensitivity; where logprobs are available, margin AUC plus a fitted threshold separates the terms.

\paragraph{For the training story.}
Preference training is not the source of the offset: the \emph{base} 0.6B model already shows it (margin AUC \auc{0.91}, $+2.6\sigma$, near-all-YES in completion format), while the base 4B is already calibrated ($-0.2\sigma$, 91.4\%). Post-training neither creates nor cures it; whether it amplifies it at small scale is format-confounded (the tuned 0.6B collapses under the base prompt, margin AUC \auc{0.60}; the tuned 4B is robust, margin AUC \auc{0.981}). The offset is a small-model pretraining property that scale removes and format perturbs; its fine-tuning dynamics remain open.

\paragraph{Limitations.}
Our domains are synthetic and English; whether the ladder's ordering holds on natural, noisy-label tasks is untested. Models are $\le$8B. Probes remain correlational: we do not causally steer the threshold at the activation level. Part of the logic-domain signal is candidate-intrinsic; our strongest per-family claims rest on the cells where that channel is silenced, and on the maze domain that closes it. Minimal foils are LLM-generated (rubric-constrained, independently checked); the deterministic families remove this dependence.

\paragraph{Conclusion.}
When do internal probes beat reading the answer? In our generation setting, not demonstrably: surface and label artifacts reproduce the published pattern. In verification, a premise-dependent validity signal forms mid-network, reaches the model's own logits, and is lost at one miscalibrated threshold that one fitted parameter repairs. Measured against matched baselines, the gap between knowledge and expression is narrower, later, and far more fixable than ``hidden knowledge'' suggests.

\bibliography{references}

@inproceedings{orgad2025llms,
  title     = {{LLM}s Know More Than They Show: On the Intrinsic Representation of {LLM} Hallucinations},
  author    = {Orgad, Hadas and Toker, Michael and Gekhman, Zorik and Reichart, Roi and Szpektor, Idan and Kotek, Hadas and Belinkov, Yonatan},
  booktitle = {Proceedings of the Thirteenth International Conference on Learning Representations (ICLR)},
  year      = {2025}
}

@inproceedings{azaria2023internal,
  title     = {The Internal State of an {LLM} Knows When It's Lying},
  author    = {Azaria, Amos and Mitchell, Tom},
  booktitle = {Findings of the Association for Computational Linguistics: EMNLP 2023},
  pages     = {967--976},
  year      = {2023}
}

@misc{kadavath2022language,
  title         = {Language Models (Mostly) Know What They Know},
  author        = {Kadavath, Saurav and Conerly, Tom and Askell, Amanda and Henighan, Tom and Drain, Dawn and Perez, Ethan and Schiefer, Nicholas and Hatfield-Dodds, Zac and DasSarma, Nova and Tran-Johnson, Eli and others},
  year          = {2022},
  eprint        = {2207.05221},
  archivePrefix = {arXiv}
}

@inproceedings{burns2023discovering,
  title     = {Discovering Latent Knowledge in Language Models Without Supervision},
  author    = {Burns, Collin and Ye, Haotian and Klein, Dan and Steinhardt, Jacob},
  booktitle = {Proceedings of the Eleventh International Conference on Learning Representations (ICLR)},
  year      = {2023}
}

@inproceedings{marks2024geometry,
  title     = {The Geometry of Truth: Emergent Linear Structure in Large Language Model Representations of True/False Datasets},
  author    = {Marks, Samuel and Tegmark, Max},
  booktitle = {First Conference on Language Modeling (COLM)},
  year      = {2024}
}

@inproceedings{li2023inference,
  title     = {Inference-Time Intervention: Eliciting Truthful Answers from a Language Model},
  author    = {Li, Kenneth and Patel, Oam and Vi{\'e}gas, Fernanda and Pfister, Hanspeter and Wattenberg, Martin},
  booktitle = {Advances in Neural Information Processing Systems 36 (NeurIPS)},
  year      = {2023}
}

@inproceedings{zhou2026geometry,
  title     = {The Geometry of Reasoning: Flowing Logics in Representation Space},
  author    = {Zhou, Yufa and Wang, Yixiao and Yin, Xunjian and Zhou, Shuyan and Zhang, Anru R.},
  booktitle = {Proceedings of the Fourteenth International Conference on Learning Representations (ICLR)},
  year      = {2026}
}

@article{belinkov2022probing,
  title     = {Probing Classifiers: Promises, Shortcomings, and Advances},
  author    = {Belinkov, Yonatan},
  journal   = {Computational Linguistics},
  volume    = {48},
  number    = {1},
  pages     = {207--219},
  year      = {2022}
}

@inproceedings{hewitt2019designing,
  title     = {Designing and Interpreting Probes with Control Tasks},
  author    = {Hewitt, John and Liang, Percy},
  booktitle = {Proceedings of the 2019 Conference on Empirical Methods in Natural Language Processing (EMNLP)},
  pages     = {2733--2743},
  year      = {2019}
}

@misc{nostalgebraist2020logit,
  title        = {Interpreting {GPT}: The Logit Lens},
  author       = {nostalgebraist},
  year         = {2020},
  howpublished = {\url{https://www.lesswrong.com/posts/AcKRB8wDpdaN6v6ru/interpreting-gpt-the-logit-lens}},
  note         = {Accessed: 2026-07-20}
}

@misc{belrose2023eliciting,
  title         = {Eliciting Latent Predictions from Transformers with the Tuned Lens},
  author        = {Belrose, Nora and Furman, Zach and Smith, Logan and Halawi, Danny and Ostrovsky, Igor and McKinney, Lev and Biderman, Stella and Steinhardt, Jacob},
  year          = {2023},
  eprint        = {2303.08112},
  archivePrefix = {arXiv}
}

@article{farquhar2024detecting,
  title     = {Detecting Hallucinations in Large Language Models Using Semantic Entropy},
  author    = {Farquhar, Sebastian and Kossen, Jannik and Kuhn, Lorenz and Gal, Yarin},
  journal   = {Nature},
  volume    = {630},
  pages     = {625--630},
  year      = {2024}
}

@misc{kossen2024semantic,
  title         = {Semantic Entropy Probes: Robust and Cheap Hallucination Detection in {LLM}s},
  author        = {Kossen, Jannik and Han, Jiatong and Razzak, Muhammed and Schut, Lisa and Malik, Shreshth and Gal, Yarin},
  year          = {2024},
  eprint        = {2406.15927},
  archivePrefix = {arXiv}
}

@inproceedings{guo2017calibration,
  title     = {On Calibration of Modern Neural Networks},
  author    = {Guo, Chuan and Pleiss, Geoff and Sun, Yu and Weinberger, Kilian Q.},
  booktitle = {Proceedings of the 34th International Conference on Machine Learning (ICML)},
  pages     = {1321--1330},
  year      = {2017}
}

@inproceedings{sharma2023towards,
  title     = {Towards Understanding Sycophancy in Language Models},
  author    = {Sharma, Mrinank and Tong, Meg and Korbak, Tomasz and Duvenaud, David and Askell, Amanda and Bowman, Samuel R. and Cheng, Newton and Durmus, Esin and Hatfield-Dodds, Zac and Johnston, Scott R. and others},
  booktitle = {Proceedings of the Twelfth International Conference on Learning Representations (ICLR)},
  year      = {2024}
}

@inproceedings{brown2020language,
  title     = {Language Models are Few-Shot Learners},
  author    = {Brown, Tom and Mann, Benjamin and Ryder, Nick and Subbiah, Melanie and Kaplan, Jared D. and Dhariwal, Prafulla and Neelakantan, Arvind and Shyam, Pranav and Sastry, Girish and Askell, Amanda and others},
  booktitle = {Advances in Neural Information Processing Systems 33 (NeurIPS)},
  pages     = {1877--1901},
  year      = {2020}
}

@inproceedings{wei2022chain,
  title     = {Chain-of-Thought Prompting Elicits Reasoning in Large Language Models},
  author    = {Wei, Jason and Wang, Xuezhi and Schuurmans, Dale and Bosma, Maarten and Ichter, Brian and Xia, Fei and Chi, Ed and Le, Quoc V. and Zhou, Denny},
  booktitle = {Advances in Neural Information Processing Systems 35 (NeurIPS)},
  pages     = {24824--24837},
  year      = {2022}
}

@inproceedings{xu2020theory,
  title     = {A Theory of Usable Information under Computational Constraints},
  author    = {Xu, Yilun and Zhao, Shengjia and Song, Jiaming and Stewart, Russell and Ermon, Stefano},
  booktitle = {Proceedings of the Eighth International Conference on Learning Representations (ICLR)},
  year      = {2020}
}

@misc{yang2025qwen3,
  title         = {Qwen3 Technical Report},
  author        = {Yang, An and Li, Anfeng and Yang, Baosong and Zhang, Beichen and Hui, Binyuan and Zheng, Bo and Yu, Bowen and Gao, Chang and Huang, Chengen and Lv, Chenxu and others},
  year          = {2025},
  eprint        = {2505.09388},
  archivePrefix = {arXiv}
}

@misc{abdin2024phi,
  title         = {Phi-3 Technical Report: A Highly Capable Language Model Locally on Your Phone},
  author        = {Abdin, Marah and Aneja, Jyoti and Awadalla, Hany and Awadallah, Ahmed and Awan, Ammar Ahmad and Bach, Nguyen and Bahree, Amit and Bakhtiari, Arash and Bao, Jianmin and Behl, Harkirat and others},
  year          = {2024},
  eprint        = {2404.14219},
  archivePrefix = {arXiv}
}

@inproceedings{xiao2023cpack,
  title     = {C-Pack: Packed Resources for General Chinese Embeddings},
  author    = {Xiao, Shitao and Liu, Zheng and Zhang, Peitian and Muennighoff, Niklas and Lian, Defu and Nie, Jian-Yun},
  booktitle = {Proceedings of the 47th International ACM SIGIR Conference on Research and Development in Information Retrieval},
  pages     = {641--649},
  year      = {2024}
}

@article{blackwell1953equivalent,
  title     = {Equivalent Comparisons of Experiments},
  author    = {Blackwell, David},
  journal   = {The Annals of Mathematical Statistics},
  volume    = {24},
  number    = {2},
  pages     = {265--272},
  year      = {1953}
}

@article{allal2025smollm2,
  title     = {SmolLM2: When Smol Goes Big -- Data-Centric Training of a Small Language Model},
  author    = {Allal, Loubna Ben and Lozhkov, Anton and Bakouch, Elie and Bl{\'a}zquez, Gabriel Mart{\'i}n and Penedo, Guilherme and Tunstall, Lewis and Marafioti, Andr{\'e}s and Kydl{\'i}{\v{c}}ek, Hynek and Lajar{\'i}n, Agust{\'i}n Piqueres and Srivastav, Vaibhav and others},
  journal   = {arXiv preprint arXiv:2502.02737},
  year      = {2025}
}

@inproceedings{zhao2021calibrate,
  title     = {Calibrate Before Use: Improving Few-Shot Performance of Language Models},
  author    = {Zhao, Zihao and Wallace, Eric and Feng, Shi and Klein, Dan and Singh, Sameer},
  booktitle = {Proceedings of the 38th International Conference on Machine Learning (ICML)},
  pages     = {12697--12706},
  year      = {2021}
}

@inproceedings{holtzman2021surface,
  title     = {Surface Form Competition: Why the Highest Probability Answer Isn't Always Right},
  author    = {Holtzman, Ari and West, Peter and Shwartz, Vered and Choi, Yejin and Zettlemoyer, Luke},
  booktitle = {Proceedings of the 2021 Conference on Empirical Methods in Natural Language Processing (EMNLP)},
  pages     = {7038--7051},
  year      = {2021}
}

@inproceedings{zheng2024large,
  title     = {Large Language Models Are Not Robust Multiple Choice Selectors},
  author    = {Zheng, Chujie and Zhou, Hao and Meng, Fandong and Zhou, Jie and Huang, Minlie},
  booktitle = {Proceedings of the Twelfth International Conference on Learning Representations (ICLR)},
  year      = {2024}
}

@inproceedings{gekhman2025inside,
  title     = {Inside-Out: Hidden Factual Knowledge in {LLM}s},
  author    = {Gekhman, Zorik and Ben-David, Eyal and Orgad, Hadas and Ofek, Eran and Belinkov, Yonatan and Szpektor, Idan and Herzig, Jonathan and Reichart, Roi},
  booktitle = {Second Conference on Language Modeling (COLM)},
  year      = {2025}
}

@inproceedings{zhao2024first,
  title     = {The First to Know: How Token Distributions Reveal Hidden Knowledge in Large Vision-Language Models?},
  author    = {Zhao, Qinyu and Xu, Ming and Gupta, Kartik and Asthana, Akshay and Zheng, Liang and Gould, Stephen},
  booktitle = {European Conference on Computer Vision (ECCV)},
  year      = {2024}
}

@inproceedings{reif2024beyond,
  title     = {Beyond Performance: Quantifying and Mitigating Label Bias in {LLM}s},
  author    = {Reif, Yuval and Schwartz, Roy},
  booktitle = {Proceedings of the 2024 Conference of the North American Chapter of the Association for Computational Linguistics: Human Language Technologies (NAACL)},
  year      = {2024}
}

@misc{park2025bridging,
  title         = {Bridging the Knowledge-Prediction Gap in {LLM}s on Multiple-Choice Questions},
  author        = {Park, Yoonah and Pyun, Haesung and Jo, Yohan},
  year          = {2025},
  eprint        = {2509.23782},
  archivePrefix = {arXiv}
}

\end{document}